\documentclass[10pt, a4paper]{article}

\usepackage[final]{lrec2026} 
\usepackage{booktabs}
\usepackage{adjustbox}
\usepackage{todonotes}
\usepackage{multicol}
\usepackage{subfigure}
\usepackage{amsmath}
\usepackage{multirow}

\makeatletter
\newcommand{\blankfootnote}[1]{%
  \begingroup
  \renewcommand{\thefootnote}{}%
  \renewcommand{\@makefnmark}{}%
  \long\def\@makefntext##1{%
    \noindent ##1%
  }%
  \footnotetext{#1}%
  \endgroup
}
\makeatother

\title{Do Language Models Encode Semantic Relations? Probing and Sparse Feature Analysis}

\name{Andor Diera, Ansgar Scherp} 

\address{Ulm University \\
         Ulm, Germany \\
         \{firstname.lastname\}@uni-ulm.de\\}

\abstract{
Understanding whether large language models (LLMs) capture structured meaning requires examining how they represent concept relationships. In this work, we study three models of increasing scale: Pythia-70M, GPT-2, and Llama 3.1 8B, focusing on four semantic relations: synonymy, antonymy, hypernymy, and hyponymy. We combine linear probing with mechanistic interpretability techniques, including sparse autoencoders (SAE) and activation patching, to identify where these relations are encoded and how specific features contribute to their representation. Our results reveal a directional asymmetry in hierarchical relations: hypernymy is encoded redundantly and resists suppression, while hyponymy relies on compact features that are more easily disrupted by ablation. More broadly, relation signals are diffuse but exhibit stable profiles: they peak in the mid-layers and are stronger in post-residual/MLP pathways than in attention. Difficulty is consistent across models (antonymy easiest, synonymy hardest). Probe-level causality is capacity-dependent: on Llama 3.1, SAE-guided patching reliably shifts these signals, whereas on smaller models the shifts are weak or unstable. Our results clarify where and how reliably semantic relations are represented inside LLMs, and provide a reproducible framework for relating sparse features to probe-level causal evidence.
\\ \newline \Keywords{semantic relations, interpretability, language models} }

\begin{document}

\maketitleabstract
\blankfootnote{The source code and datasets are available at: \href{https://github.com/drndr/semantic\_relation\_interpret}{github.com/drndr/semantic\_relation\_interpret}}
\section{Introduction}

Understanding the internal mechanisms by which large language models (LLMs) process and represent knowledge is fundamental to advancing both theoretical insights and practical applications in natural language processing (NLP). Interpretability research addresses this need by providing powerful frameworks for dissecting internal representations, moving beyond black-box evaluations to examine how specific activation patterns emerge in the models~\cite{sharkey2025open}. This deeper understanding serves two main objectives: identifying hidden biases embedded within learned representations and developing more precise methods for controlling model behavior and outputs~\cite{liu2024devil, soo2025interpretable}.

While existing interpretability research has made significant progress in identifying individual concepts and feature representations within LLMs~\cite{cunningham2023sparse, gurneefinding}, the investigation of semantic relationships between these concepts remains largely unexplored.
Certain relationships, such as synonymy and antonymy, which test whether models can distinguish similarity from opposition in meaning, and hierarchical relations like hypernymy and hyponymy, are particularly diagnostic of whether models develop structured conceptual understanding or rely on surface-level associations. Existing research suggests that LLMs still face substantial challenges in accurately predicting or encoding such relationships~\cite{roussinov2023fine, moskvoretskii2024large, cao2024comprehensive}. This limitation underscores a broader gap in our understanding of how (or if) LLMs encode relational structure, emphasizing the need for interpretability research to analyze the semantic relationships embedded within model representations.

In this paper, we ask whether semantic relation-signals exist in LLMs, where they are localized across layers/components, and how strongly they can be manipulated. To address these questions, we focus on four semantic relations: synonymy, antonymy, hypernymy, and hyponymy. We examine the Pythia-70M, GPT-2-124M, and Llama 3.1-8B models, using linear probing, sparse autoencoders (SAEs), and activation patching. Our results suggest there are no unique ``relation layers''; depth profiles are diffuse at middle layers and peak locations are unstable, yet in Llama 3.1 SAE patching reliably induces probe-level causal shifts. 
Relations differ in profile, spanning clear gaps in probe accuracy, intervention sensitivity, and the relative weight of sufficiency versus necessity. 
Antonymy is reliably captured by probes and can be causally influenced by sparse features. Synonymy is the hardest to decode and is the least affected by activation patching, indicating a more diffuse encoding. Hierarchical relations diverge: hypernymy is easy to strengthen but hard to suppress; hyponymy strengthens and collapses readily. Although our claims are limited to probe-level causality, the findings still reveal manipulability, stable difficulty ordering, and a bias in hierarchical relations.
\\
The main contributions of our paper:
\begin{itemize}
    \item A systematic comparison of how LLMs encode four semantic relations (synonymy, antonymy, hypernymy, and hyponymy) across scales from 70M to 8B parameters.
    
    \item An integrated probing and sparse feature analysis framework for identifying relation-specific activations and assessing their probe-level causal influence.
    
    \item Results show that relational information concentrates in post-residual and MLP components but remains diffusely distributed across mid-layers, with broad and uncertain depth peaks.

    \item SAE-guided activation patching isolates relation-specific features and demonstrates probe-level necessity/sufficiency, including asymmetric directionality in hierarchical relationships.

\end{itemize}

\section{Related Work}

Machine learning interpretability is often approached through two main strategies: concept-based interpretability and reverse engineering~\cite{sharkey2025open}. While both aim to make the inner workings of complex models more transparent, they differ in methodology and focus. Concept-based approaches center on aligning internal representations with human-understandable constructs, whereas reverse engineering seeks to uncover the underlying computational mechanisms without relying on predefined concepts.

\subsection{Concept-based Interpretability}

A common strategy for identifying human-interpretable concepts within a model is to train a simple classifier (often referred to as a probe) to detect their presence in intermediate representations. In the context of LLMs, probing has been used to assess encoded linguistic properties~\cite{conneau2018you, hewitt2019structural}, factual knowledge~\cite{peng2022copen, gurneelanguage}, and concept relations~\cite{aspillaga2021inspecting}.
An alternative approach represents concepts as vectors in the activation space, which enables controlled interventions by shifting activations along specific conceptual directions~\cite{bolukbasi2016man, chanin2024identifying}. Complementary to these methods, concept bottleneck models introduce an explicit intermediate layer of human-defined concepts, enabling models to make predictions through interpretable concept activations~\cite{sun2024concept}. Despite their appeal, concept-based methods face several limitations. Probes and concept vectors capture correlations rather than true causality~\cite{belinkov2022probing}, while concept bottlenecks may constrain model capacity or fail to generalize when concept sets are incomplete~\cite{yuksekgonulpost}. In general, concept-based interpretability methods impose predefined conceptual frameworks rather than uncovering the emergent features that a model has learned~\cite{sharkey2025open}.

\subsection{Mechanistic Interpretability}

In contrast to concept-based approaches, mechanistic interpretability aims to reverse-engineer the internal computation of models by analyzing circuits, neurons, and attention patterns~\cite{sharkey2025open}. In LLMs, this approach seeks to reveal not only what linguistic or factual information is present, but also how such information is processed, represented, and transformed across layers. Among the earliest demonstrations of mechanistic interpretability in transformer-based language models was the discovery of induction heads (attention heads that perform token copying), providing a functional explanation for how LLMs generalize patterns through in-context learning~\cite{elhage2021mathematical}. Subsequent research has uncovered a range of interpretable behaviors in LLMs, including mechanisms for object identification~\cite{wanginterpretability}, factual recall~\cite{nanda2023fact, geva2023dissecting}, and arithmetic functions~\cite{stolfo2023mechanistic, quirke2024increasing}.

A key challenge in mechanistic interpretability is the polysemanticity of individual neurons, where a single neuron often activates for multiple unrelated features, making it difficult to assign consistent functional roles~\cite{elhage2022toy}. One promising direction for mitigating this issue is promoting monosemanticity, where neurons are aligned with single, interpretable features~\cite{bricken2023towards}. Sparse autoencoders have emerged as a useful tool in this context, enabling the extraction of disentangled features from model activations by encouraging sparsity in the latent variables~\cite{cunningham2023sparse}.

\subsection{Semantic Relationships in LLMs}

While LLMs can represent many individual concepts, their ability to model semantic relationships remains limited and inconsistent~\cite{roussinov2023fine, moskvoretskii2024large, cao2024comprehensive}. As~\citet{roussinov2023fine} notes, pre-trained language models often appear to perform well on benchmark datasets due to lexical memorization, rather than a genuine understanding of relational structure. The study by~\citet{moskvoretskii2024large} further shows that LLMs predict hypernym relations more successfully than hyponyms, suggesting an asymmetry in how relational information is encoded. Meanwhile, a comprehensive evaluation by~\citet{cao2024comprehensive} finds that models continue to lag significantly behind humans across nearly all types of semantic relation prediction tasks.

One persistent source of error is the distributional paradox of antonyms: because antonyms often share contexts (e.\,g., happy and sad with person, feeling, day), they appear highly similar in the distributional space despite opposing meanings~\cite{cruse1986lexical}. This phenomenon complicates the separation of antonymy from synonymy in vector spaces and has been observed across both static and contextual embeddings~\cite{scheible2013uncovering,fodor2023importance}. The paradox illustrates a broader issue: LLMs capture co-occurrence regularities but often fail to abstract underlying semantic relations.

From an interpretability perspective, relatively little is known about how such relations are internally represented.~\citet{aspillaga2021inspecting} extracted concept knowledge graphs from language models using probing classifiers, but found several imprecisions in the organization of concepts. \citet{park2024geometry} further investigated this by analyzing the geometry of categorical and hierarchical concepts in LLMs, finding that categories form polytopes (e.\,g., simplices) and that hierarchical relationships correspond to orthogonality between their vector representations in the embedding space. 

\section{Procedure}

\subsection{Models}

For our analysis, we use three decoder-only transformer language models. 
Pythia-70M~\cite{biderman2023pythia} is a 6-layer open-weight model developed by EleutherAI to support mechanistic interpretability research. GPT-2~\cite{radford2019language} is a 12-layer model with 124M parameters that has become a standard reference point in interpretability work. Finally, Llama 3.1 8B~\cite{grattafiori2024llama} is a 32-layer model from a recent family of open-weight LLMs that enables larger-scale analysis and serves as a strong baseline for evaluating the scaling of our approach. 
For experiments with sparse autoencoders, we utilize model- and layer-matched pre-trained SAE models provided by the SAELens library~\cite{bloom2024saetrainingcodebase}\footnote{The SAELens dictionary IDs used for each model and layer are specified in the accompanying code repository}.

\subsection{Datasets}

For interpreting natural language concepts, we use WordNet~\citelanguageresource{lr_wordnet}, a large lexical database of English that organizes words into sets of synonyms called synsets, each representing a distinct concept. Synsets are interlinked through a network of semantic relationships such as hypernyms (more general terms), hyponyms (more specific terms), synonyms (words with similar meanings), and antonyms (words with opposite meanings).

\begin{table}[h!]
    \small
    \centering
    \begin{tabular}{l|l|l}
        \toprule
        \textbf{Relation} & \textbf{Description}        & \textbf{Example} \\
        \midrule
        Synonym & Same meaning & happy $\rightarrow$ joyful \\
        Antonym & Opposite meaning & happy $\rightarrow$ sad \\
        Hypernym   & More general       & dog $\rightarrow$ animal \\
        Hyponym    & More specific      & dog $\rightarrow$ beagle \\
        Random & No semantic relation & dog $\rightarrow$ table\\
        \bottomrule
        \end{tabular}
        \caption{Overview of the WordNet-derived relation dataset. Each example pair illustrates a semantic relation used in our experiments.}
\end{table}

We constructed our dataset by querying WordNet~3.0 via the NLTK interface~\cite{bird2006nltk}. To ensure high lexical quality, we restricted candidates to words longer than two characters, fully alphabetic, lowercase, and without underscores or hyphenation. From WordNet synsets, we randomly extracted 1,000 pairs for each semantic relation: hypernyms, hyponyms, synonyms, antonyms, along with 1,000 random word pairs that were verified to share no semantic relationship. 
Duplicate entries were removed, and hypernym and hyponym sets were constructed to be unidirectional with no overlap. Furthermore, to maintain lexical balance, we ensure that the dataset preserves the distribution of the parts of speech balance in the full WordNet dataset, consisting of 66\% nouns, 23\% verbs, and 11\% adjectives. The resulting word pairs are split into train/test with a stratified 80/20 ratio, enforcing lemma-disjoint splits to prevent overlap between sets. After splitting, we augmented each pair with three neutral contextual prompts (\textit{"The word {A} relates to {B}"}, \textit{"{A} and {B} are connected"}, \textit{"Consider {A} and {B} together"}), yielding a total of 15,000 instances.

\subsection{Probing}
We probe the internal representations of each model to assess how well concept relationships are captured across layers. From every transformer block, we extract three activation streams: the attention output, the MLP output, and the post-residual stream. We apply mean pooling across all tokens to obtain a single sequence representation for each layer and train a multinomial logistic regression probe, using train-only per-feature z-scoring ($(\mu,\sigma;\ \sigma \ge 10^{-6})$) and fixed specifications. Probes are fitted on the training split and evaluated on a held-out test split.

Performance is summarized with four metrics: mean accuracy across layers (average test accuracy over all layers), peak accuracy (maximum test accuracy across layers), peak depth (layer index at which accuracy is maximal), and center of mass, which captures how representational strength is distributed across depth. The center of mass is defined as the accuracy-weighted average layer index,
\[
\mathrm{CoM} = \frac{\sum_{l=0}^{L-1} l \, a_l}{\sum_{l=0}^{L-1} a_l},
\]
where $a_l$ denotes the probe's test accuracy at layer $l$. For cross-model interpretability, we normalize the center of mass and peak depth to [0,1] by dividing their layer indices by the maximum possible depth (e.\,g. L-1). Uncertainty is quantified with 95\% confidence intervals, obtained by stratified bootstrap over the test set: we resample within classes with replacement, recompute the metrics for each replicate, and take the 2.5th–97.5th percentiles~\cite{bestgen2022please}.

\subsection{Activation Patching}
\label{sec:patching}

To examine whether relation-specific SAE features causally affect a linear probe's decision over the post-residual stream, we use activation patching~\cite{meng2022locating}, 
a targeted intervention technique that replaces or removes internal activations to test their effect on output logits. In our setup, each relation type is captured by a small subset of SAE features with the largest linear-probe coefficients. For each model/layer, we train the probe on the training split, rank features by the absolute coefficient magnitude, and choose $k$ on a held-out validation set. To keep interventions sparse and interpretable, we cap $k$ at 1\% of the SAE dictionary ($d = 32,768$). This mirrors common SAE practice of reporting results at small fixed top-$k$ budgets and operating on only a tiny fraction of latents per example~\cite{gaoscaling}. We then intervene on the selected features in the test set by either ablating them (setting their activations to zero) or injecting them into neutral, relation-free inputs.

Building on the activation patching framework of~\citet{zhangtowards}, we define two causal measures adapted to our SAE probe setup. Following the general distinction in causal interpretability between testing whether a component is required for a behavior (necessity) and whether it alone can produce it (sufficiency)~\cite{heimersheim2024use}, we operationalize both notions in terms of the probe's semantic logit difference. Necessity quantifies the loss of relation evidence when the associated SAE features are ablated, whereas sufficiency measures the gain of evidence when the same features are injected into neutral, relation-free inputs.

\paragraph{Semantic Logit Difference.}
To provide a unified causal measure for both necessity and sufficiency, 
we define a semantic logit difference ($\text{LD}_{\text{sem}}$) that captures how strongly the probe favors the target relation 
over its strongest semantic rival.  Rather than using raw class probabilities, we measure the margin between the target relation and its closest semantic competitor. A larger margin means the probe is more certain about its relational judgment. Formally, 
\[
\text{LD}_{\text{sem}} = \text{logit}_{t} - \max_{c \notin \{t,\text{random}\}} \text{logit}_{c},
\]
where $t$ denotes the target relation class. 
This margin-based metric isolates the model’s preference for the target relation while excluding the \emph{random} (no-relation) class.

\paragraph{Sufficiency}
The goal of sufficiency is to test whether the selected SAE features alone can create a relational signal where none existed.
We define it as the extent to which injecting a relation’s top-$k$ SAE features into neutral examples increases the probe’s preference for that relation. 
Let $x$ denote a representation from the \emph{random} class, and $x^{\text{patch}}$ the same representation after patching with the target’s top-$k$ features. 
Sufficiency is quantified as the mean change in semantic logit difference,
\[
\Delta \text{LD}_{\text{sem}} = \text{LD}_{\text{sem}}(x^{\text{patch}}) - \text{LD}_{\text{sem}}(x),
\]
where positive values indicate that the patched representation more strongly favors the target relation.
In addition, we report the change in flip-rate ($\Delta\text{FR}$) we as a behavioral measure,
\[
\Delta\text{FR} = \Pr[\hat{y}_{\text{sem}}(x^{\text{patch}})=t] - \Pr[\hat{y}_{\text{sem}}(x)=t],
\]
where $\hat{y}_{\text{sem}}$ denotes the probe’s predicted relation label when restricted to the set of semantic classes 
(i.e., excluding the \emph{random} class). 
The $\Delta$~FR thus captures the net proportion of examples whose predicted relation switches toward the target after patching ($\mathrm{FR} \in [-1,1]$), 
with higher values reflecting greater causal sufficiency of the injected features.

\paragraph{Necessity.}
While sufficiency injects features into neutral inputs, necessity removes them from inputs where the relation is already present, testing whether the signal collapses without them. We define necessity as the extent to which ablating a relation’s top-$k$ SAE features reduces the probe’s preference for that relation. 
Let $x$ be an example from relation $t$ (with $t \neq \text{\emph{random}}$), and $x^{\text{abl}}$ the same example after ablating (scaling to 0) the corresponding top-$k$ SAE features. 
Necessity is quantified as the mean change in semantic logit difference,
\[
\Delta \text{LD}_{\text{sem}} \;=\; \text{LD}_{\text{sem}}(x^{\text{abl}})\;-\;\text{LD}_{\text{sem}}(x),
\]
where more negative values indicate greater necessity of the ablated features. 
As a behavioral counterpart, we report the drop rate (DR),
\[
\text{DR} \;=\; \Pr\!\big[\hat{y}_{\text{sem}}(x)=t \;\wedge\; \hat{y}_{\text{sem}}(x^{\text{abl}})\neq t\big],
\]
where $\hat{y}_{\text{sem}}$ is the probe’s prediction over semantic classes (excluding random). 
The drop rate captures the proportion of items whose predicted relation leaves the gold class after ablation ($\mathrm{DR} \in [0,1]$), reflecting the causal necessity of the ablated features.

Activation patching effects are quantified on the SAE probe margin/behavior, not on model outputs; we therefore do not claim token-prediction-level causality.

\label{sec:probing}
\section{Results}
Our results are organized in three parts. We begin with a general probing analysis across layers, blocks, and models, comparing dense residual activations. We then conduct relation-specific analyses, beginning with the contrast between antonymy and synonymy, followed by an examination of hierarchical relations (hypernymy/hyponymy). Finally, we examine sparse features directly, asking whether a small set of SAE dimensions carries the relation signal via top-$k$ feature analysis and targeted ablations.

\subsection{Probing Results}
The results of probing the internal model representations are shown in Table~\ref{tab:dense_probing}.
The table highlights a clear scale effect and a consistent difficulty ordering across models, with antonymy easiest, followed by hyponymy, then hypernymy, and synonymy hardest. Accuracies rise from Pythia to GPT-2 to Llama 3.1 in both means and peak accuracies; antonymy approaches ceiling on Llama, while synonymy remains the hardest even at scale. Hypernymy and hyponymy occupy the middle with consistent spacing. Random controls increase with scale but remain below most structured relations. Peak accuracy consistently exceeds layer means by 0.02–0.07 points, indicating mild, uneven layer-wise variation without presuming sharp localization.

\begin{figure*}[h]
    \centering
    \includegraphics[width=0.325\textwidth]{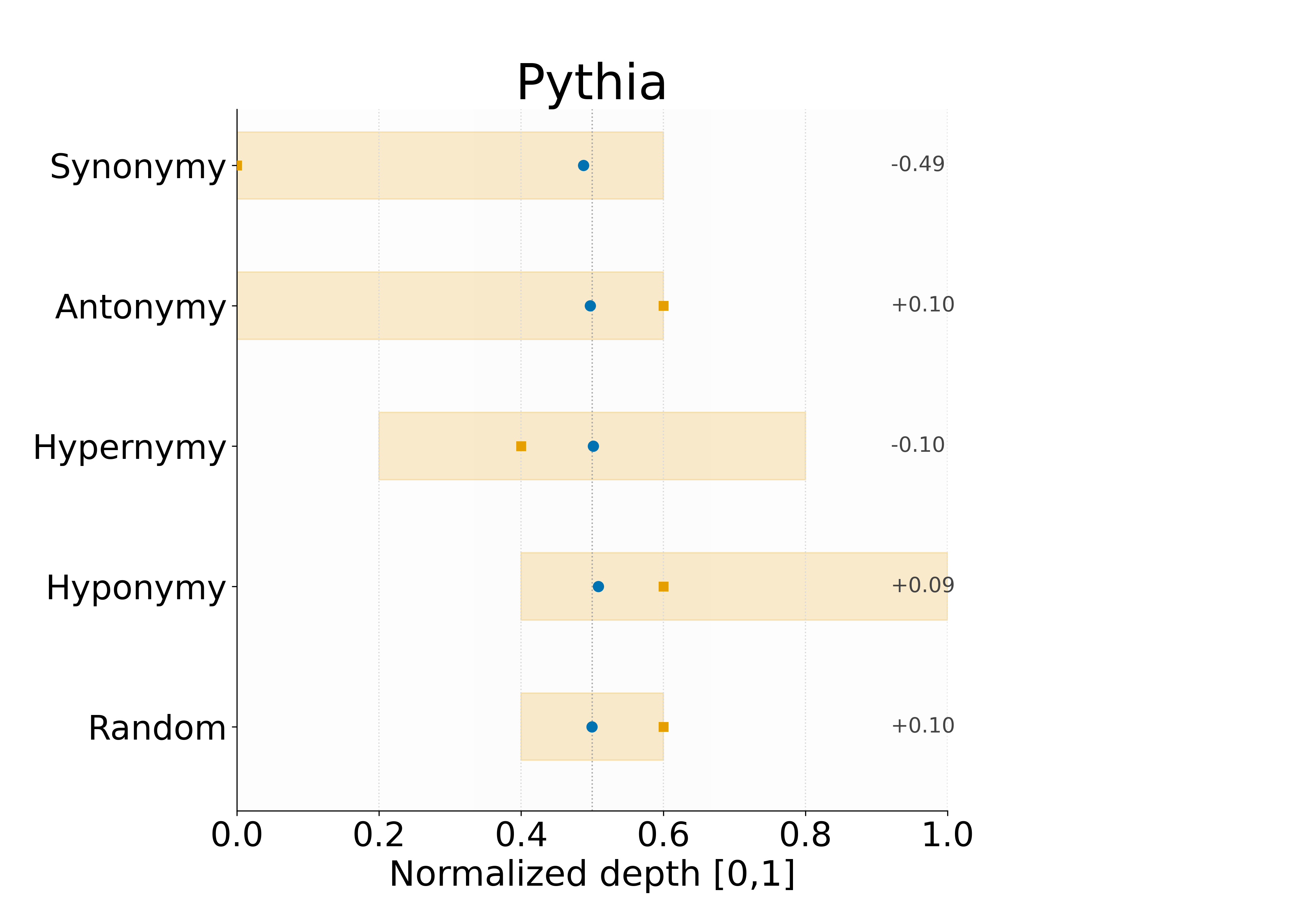}
    \hfill
    \includegraphics[width=0.325\textwidth]{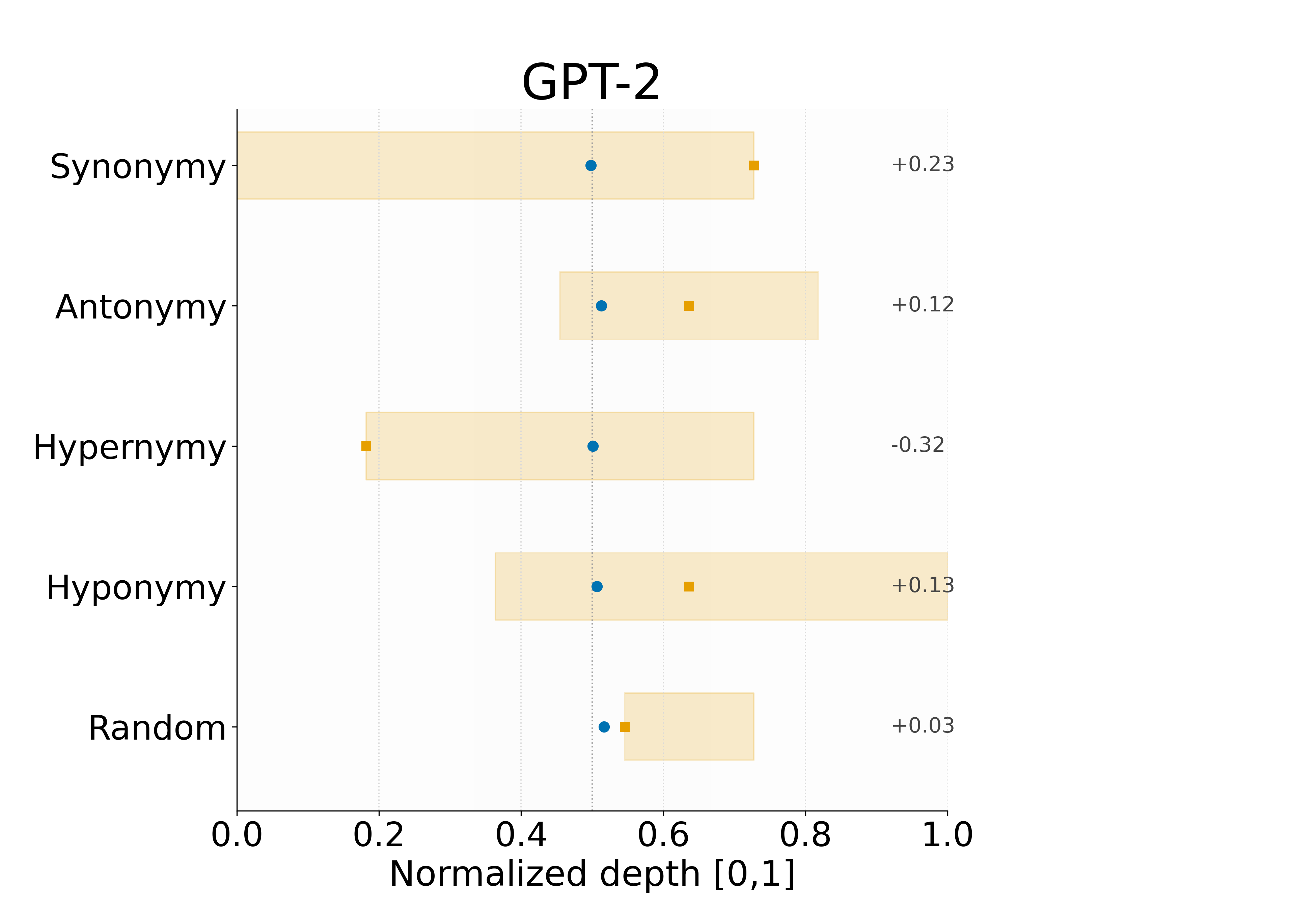}
    \hfill
    \includegraphics[width=0.325\textwidth]{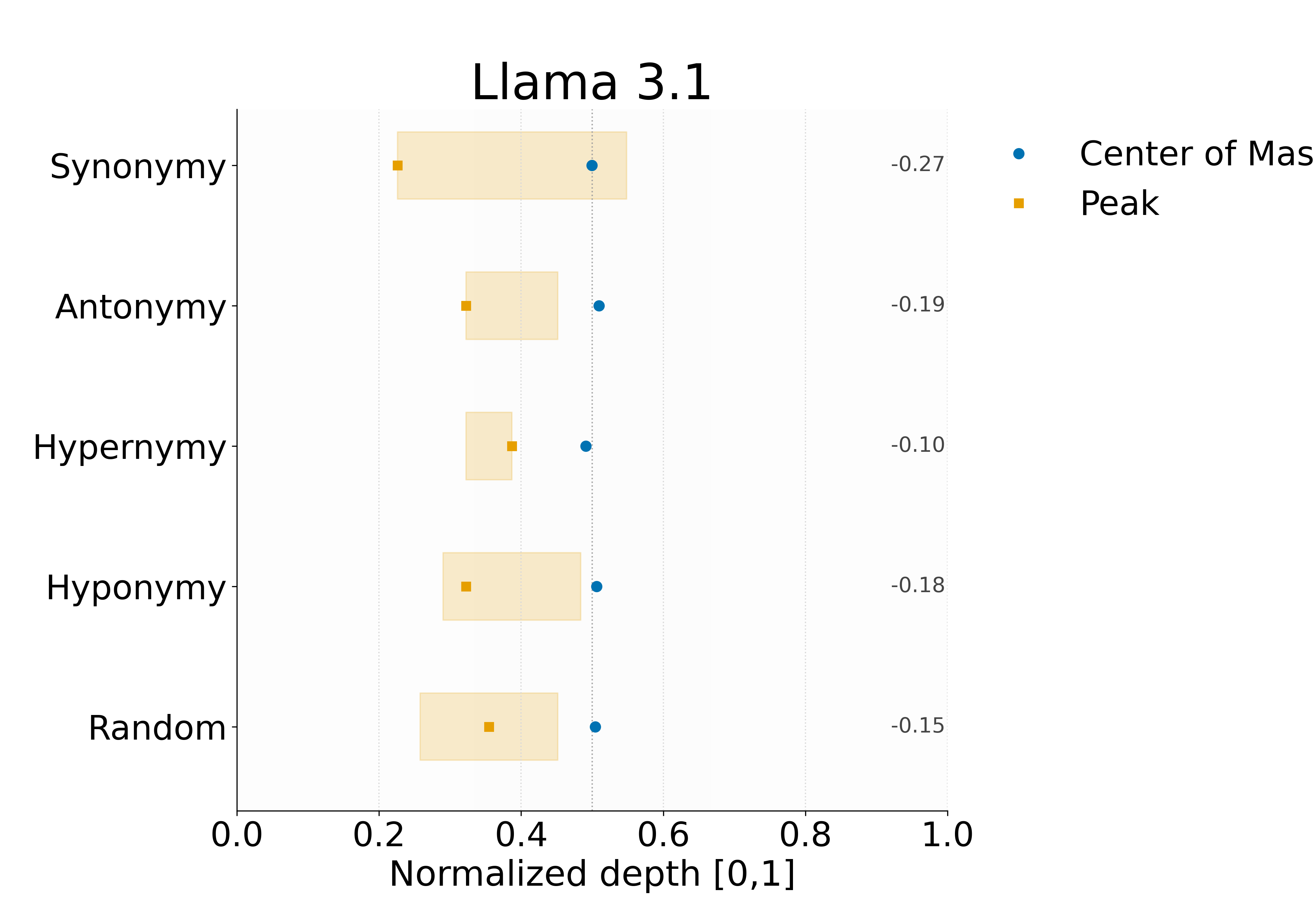}
    \caption{Probing results on dense post-residual representations in the three models. Blue circles mark the center of mass (COM) of layer accuracies; orange squares mark the peak depth. Shaded bars show the 95\% confidence interval. Numbers at right indicate Peak–COM difference.}
    \label{fig:depth}
\end{figure*}

Figure~\ref{fig:depth} shows that localization is weakly structured. Centers of mass sit near the mid‐stack (0.5) for all relations and models, while peak layers shift with no consistent ordering across relations. Hyponymy shows a slight late skew on smaller models; antonymy and hypernymy show small, model-dependent displacements. Synonymy is the least stable by its peak depth. The shaded peak-depth confidence intervals are broad overall, confirming diffuse rather than tightly localized peaks. Overall, scale increases accuracy (cf. Table~\ref{tab:dense_probing}), and while the layer of maximal evidence varies by relation and model, the bulk of the signal sits in mid layers, aligning with prior reports that lexical semantics emerge mid-stack~\cite{tenney2019bert}.

\begin{figure*}[h]
    \centering
    \includegraphics[width=0.325\textwidth]{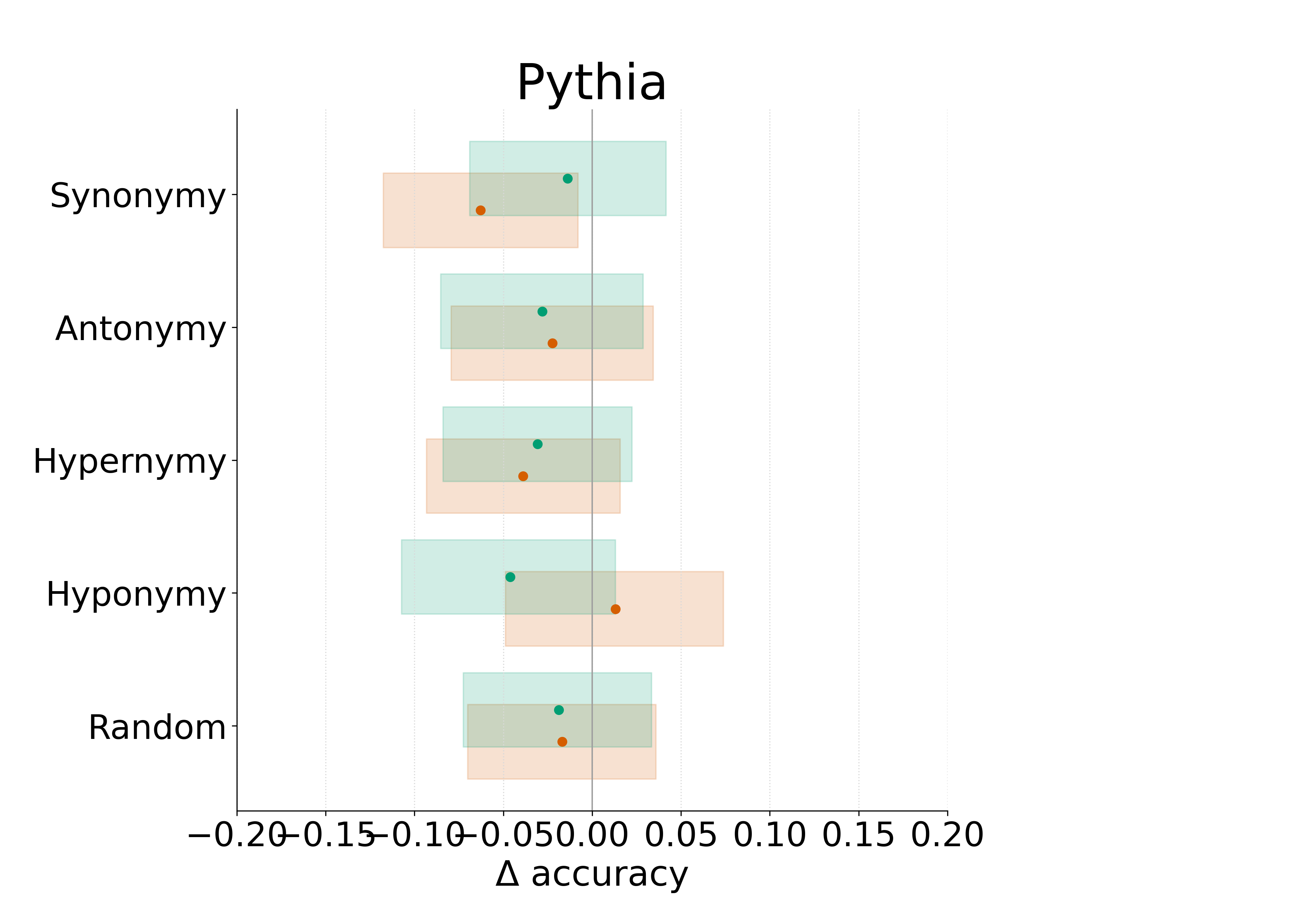}
    \hfill
    \includegraphics[width=0.325\textwidth]{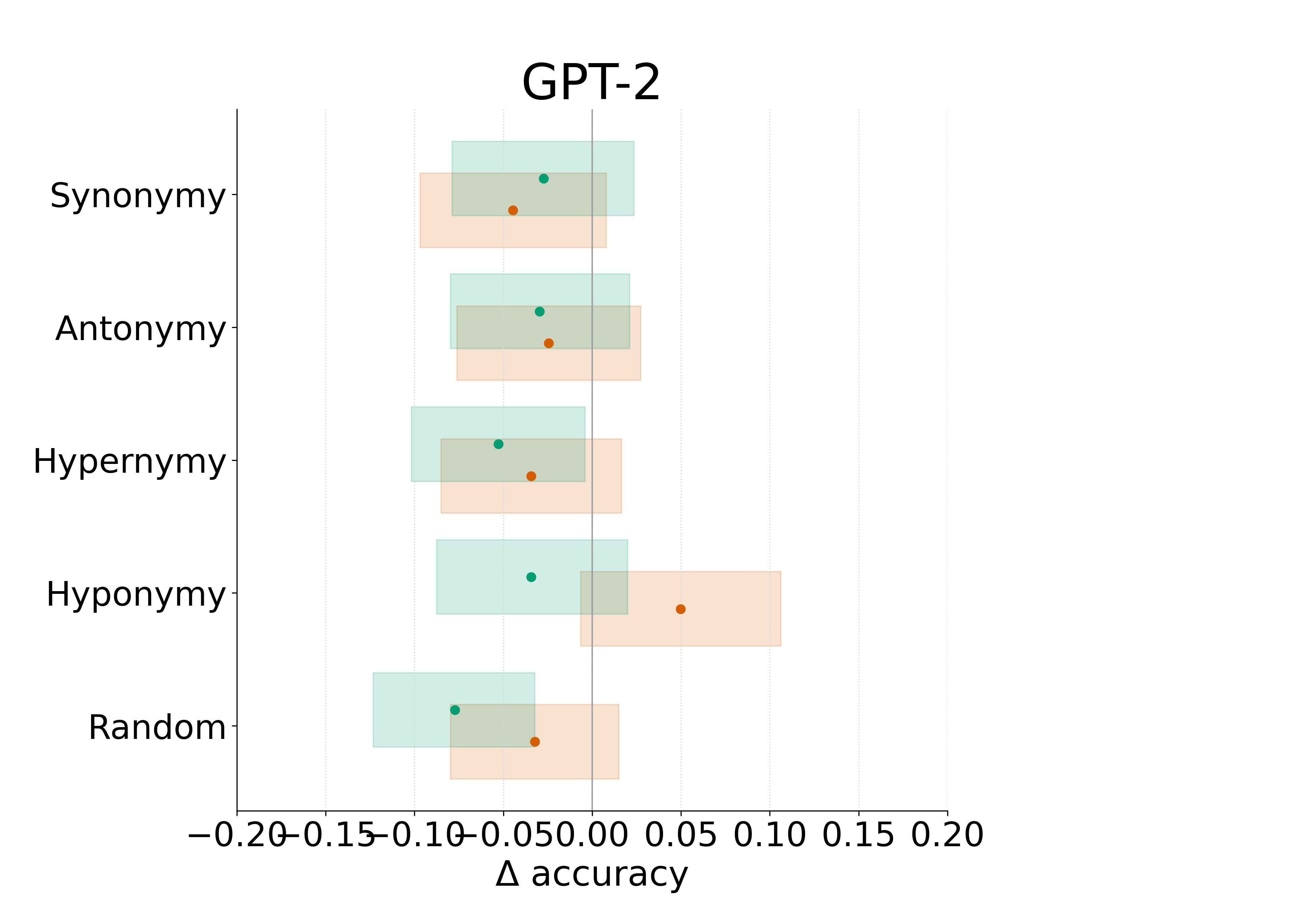}
    \hfill
    \includegraphics[width=0.325\textwidth]{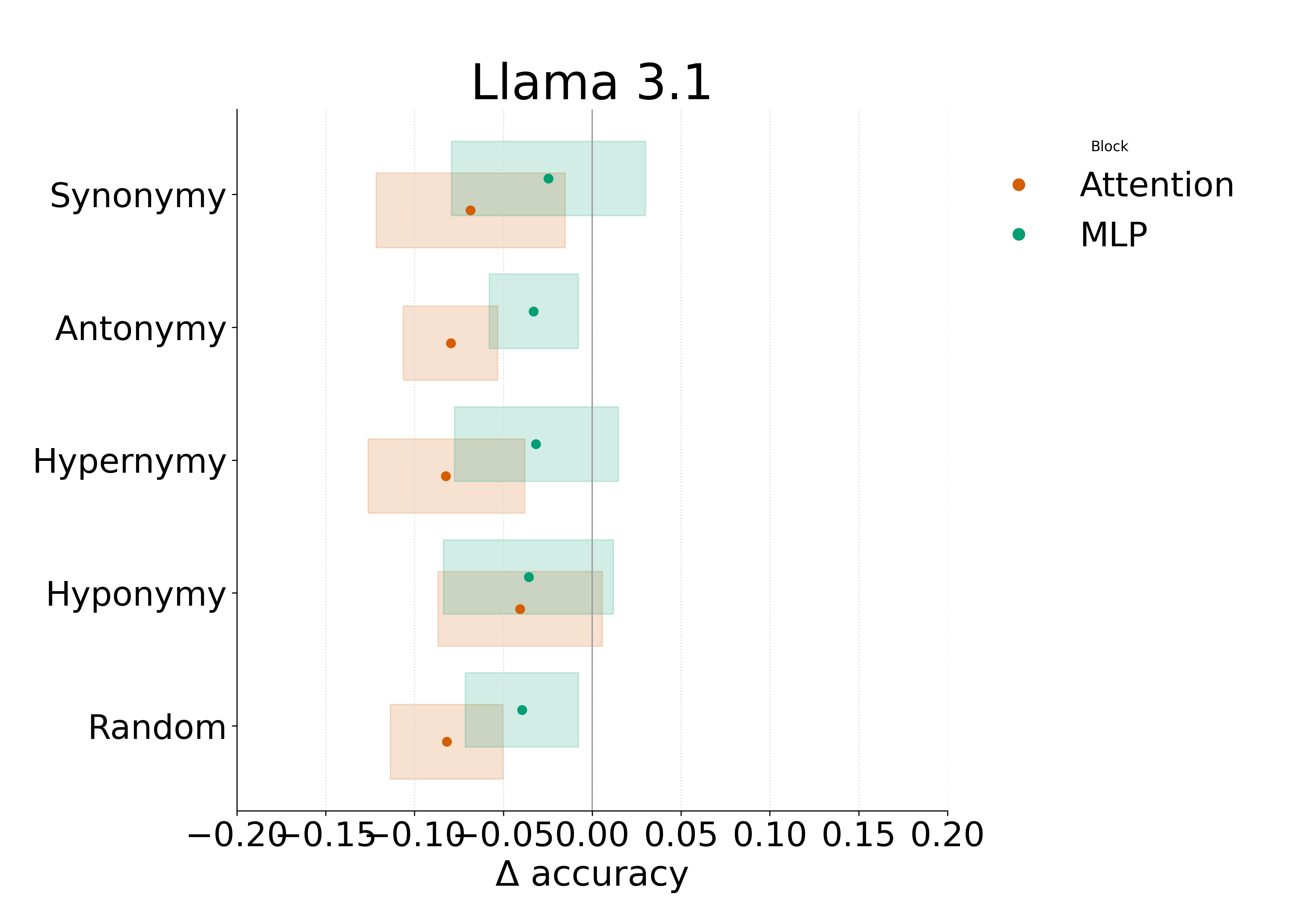}
    \caption{Block-wise accuracy differences relative to the post-residual stream. Points show mean $\Delta$ accuracy (Attention/MLP probes compared to the post-residual probe); shaded bands represent the 95\% confidence intervals.}
    \label{fig:block-comparison}
\end{figure*}

Figure~\ref{fig:block-comparison} displays the differences between the components of the transformer block, shown as the accuracy deltas of attention and MLP relative to the post-residual stream (the complete output of the block passed to the next layer). The post-residual target is the strongest overall: most deltas are below zero. MLP deltas are modestly higher than attention in Llama 3.1, but are mostly on par with it in the smaller models. Synonymy shows the largest deficits relative to the reference, hypernymy and antonymy hover around zero to mildly negative, and hyponymy is positive in the attention stream for Pythia and GPT-2 but not for Llama. Overall, the results suggest that the full-block output carries the most recoverable signal; the MLP is slightly behind; and any attention advantage is relation-specific and weakens with scale. 

\begin{table}[]
    \small
    \centering
    \begin{adjustbox}{width=0.49\textwidth}
    \begin{tabular}{l|cc|cc|cc}
        \toprule
        & \multicolumn{2}{c|}{Pythia} & \multicolumn{2}{c|}{GPT-2} & \multicolumn{2}{c}{Llama 3.1} \\
        Relation & Mean & Peak & Mean & Peak & Mean & Peak \\
        \midrule
        Syn   & 0.46$_{0.03}$ & 0.49$_{0.03}$ & 0.50$_{0.02}$ & 0.54$_{0.03}$ & 0.72$_{0.02}$ & 0.74$_{0.03}$ \\
        Ant   & \textbf{0.65}$_{0.03}$ & \textbf{0.68}$_{ 0.03}$ & \textbf{0.68}$_{0.02}$ & \textbf{0.73}$_{0.03}$ & \textbf{0.91}$_{0.01}$ & \textbf{0.98}$_{0.01}$ \\
        Hyper & 0.50$_{0.03}$ & 0.54$_{0.03}$ & 0.52$_{0.02}$ & 0.56$_{0.03}$ & 0.66$_{0.02}$ & 0.75$_{0.03}$ \\
        Hypo  & 0.55$_{0.03}$ & 0.58$_{0.03}$ & 0.59$_{0.03}$ & 0.64$_{0.03}$ & 0.80$_{0.02}$ & 0.84$_{0.02}$ \\
        Rand  & 0.49$_{0.02}$ & 0.55$_{0.03}$ & 0.62$_{0.02}$ & 0.68$_{0.03}$ & 0.86$_{0.01}$ & 0.93$_{0.01}$ \\
        \bottomrule
    \end{tabular}
    \end{adjustbox}
    \caption{Linear probing on dense representations. Mean: accuracy averaged across layers. Peak: maximum accuracy across layers for each model and relation. 95\% confidence interval reported as half-width in subscript.}
    \label{tab:dense_probing}
\end{table}

\subsection{Relation-specific Analysis}
\subsubsection{Antonymy vs. Synonymy}
From an interpretability perspective, antonymy relations expose the limits of distributional similarity. If models encode only surface co-occurrence, synonym–antonym pairs should remain as close as synonym–synonym pairs. In contrast, a structured relational representation should force a divergence. While this paradox is well documented for static embeddings~\cite{scheible2013uncovering, fodor2023importance}, contextualized models process full input sequences and could in principle leverage surrounding context to disambiguate synonyms from antonyms. To further test this, we measured cosine similarity for intra-group (Syn–Syn, Ant–Ant) and inter-group (Syn–Ant, Syn–Rand, Ant–Rand) word pairs across embedding, middle, and final layers without the contextual prompts used in the probing experiments.

\begin{figure*}[!th]
    \centering \includegraphics[width=0.98\textwidth]{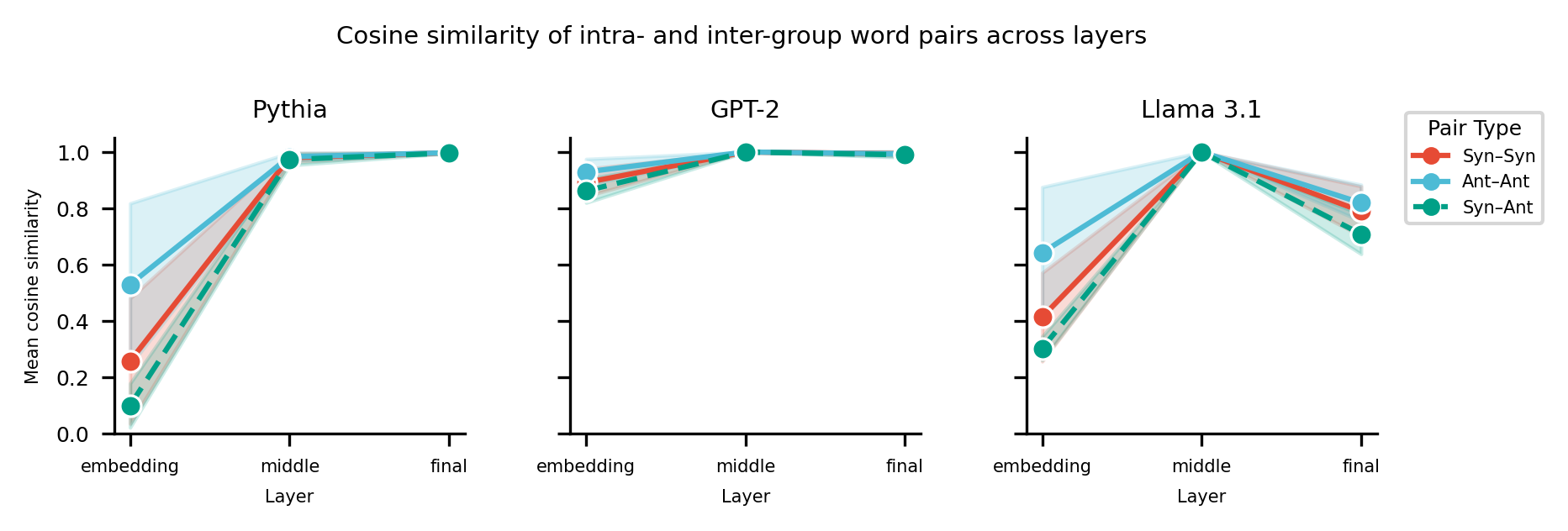}
    \caption{Average cosine similarity for intra-group (synonym–synonym, antonym–antonym) and inter-group (synonym–antonym) word pairs across embedding, middle, and final layers in three models.}
    \label{fig:cosine}
\end{figure*}

The results in Figure \ref{fig:cosine} show that cosine similarity is a poor diagnostic of relational encoding. In Pythia, modest synonym–antonym distinctions are present at the embedding layer but collapse completely afterward. GPT-2 and Llama 3.1 likewise converge toward near-identity across relations, with only a slight re-emergence at the final layer in Llama. In other words, raw closeness homogenizes representations and erases lexical contrasts, regardless of scale.

However, this collapse contrasts sharply with the probing results (Section~\ref{sec:probing}): antonymy remains linearly separable even in small models and becomes almost perfectly recoverable in Llama. Synonymy, although weaker, also emerges reliably in large models. Taken together, the two analyses suggest that distributional similarity measures are misleading: relations are not encoded as overall similarity but in specific internal activations that are not reflected in final embedding similarities. In other words, while contextualization provides the representational capacity to distinguish antonymy from synonymy, this information is not captured by similarity-based measures and can be only recovered through directed probing.

\subsubsection{Hypernymy vs. Hyponymy}
\label{sec:hyper-hypo}
Hypernymy and hyponymy offer a natural test of directionality in lexical representations. Unlike synonymy or antonymy, these relations are asymmetric: if dog is a hyponym of animal, the inverse does not hold. Any symmetric similarity measure, such as cosine, is therefore blind to the distinction. For a model to genuinely encode hierarchical lexical structure, it must represent not only that two words are related but also the direction of that relation.

To evaluate this, we construct two additional test sets from the hypernymy, hyponymy, and random pairs: one in the original order (e.\,g., ``The word animal relates to dog'' labeled as hypernymy) and one with the order reversed (``The word dog relates to animal'') and the label inverted (hyponymy). A model that truly encodes directionality should allow the probe to achieve comparable performance on both sets, since reversing the order should simply invert the class. If instead the model only encodes undirected relatedness, performance will diverge between the two, with accuracy dropping or becoming skewed when the order is reversed. We include random pairs as a control, since their label is unaffected by reversal and should yield consistent performance across both sets.

Table~\ref{tab:hyper-hypo} reports the accuracy difference ($\Delta$) between original and reversed evaluations at the peak layer for each relation type. As expected, random pairs remain stable ($\Delta < 0.5$), confirming that reversal alone does not degrade performance when no direction is involved. Hypernymy likewise shows only modest drops ($\Delta$ ranges from $0.07$ to $0.09$), suggesting that upward generalization is relatively easy to capture. Hyponymy, however, exhibits much larger gaps ($\Delta$ ranges from $0.24$ to $0.29$), indicating that performance deteriorates substantially when direction is inverted. This asymmetry is consistent across scales: even Llama retains an accuracy difference of $0.24$. We also note that peak performance often occurs at different layers for original and reversed evaluations, 
suggesting that directionality is not localized but spread unevenly across the network. Taken together, these results suggest that models encode lexical hierarchies in a biased fashion, favoring hypernymy over hyponymy and failing to represent directionality in a balanced way.

\begin{table}
    \centering
    \begin{adjustbox}{width=0.45\textwidth}
    \begin{tabular}{l l | rrr}
        \toprule
        Relation & & Pythia & GPT-2 & Llama 3.1 \\
        \midrule
        \midrule
        \multirow{3}{*}{Hyper}
            & $\text{Acc}_{\text{orig}}$ & 0.69$_{0.03}$ & 0.71$_{0.03}$  & 0.88$_{0.02}$\\
            & $\text{Acc}_{\text{flip}}$ & 0.60$_{0.03}$ & 0.63$_{0.03}$ & 0.81$_{0.02}$ \\
            & $\Delta$ & -0.09 & -0.08 & -0.07 \\
        \midrule
        \multirow{3}{*}{Hypo}
             & $\text{Acc}_{\text{orig}}$ & 0.67$_{0.03}$ & 0.74$_{0.03}$ & 0.89$_{0.02}$ \\
            & $\text{Acc}_{\text{flip}}$ & 0.43$_{0.03}$ & 0.45$_{0.03}$ & 0.65$_{0.03}$\\
            & $\Delta$ & -0.24 & -0.29 & -0.24 \\
        \midrule
        \multirow{3}{*}{Rand}
            & $\text{Acc}_{\text{orig}}$ & 0.67$_{0.03}$ & 0.75$_{0.02}$ & 0.91$_{0.01}$ \\
            & $\text{Acc}_{\text{flip}}$   & 0.64$_{0.03}$ & 0.71$_{0.02}$ & 0.91$_{0.01}$ \\
            & $\Delta$ & -0.03 & -0.04 & 0.00 \\
        \bottomrule
    \end{tabular}
    \end{adjustbox}
    \caption{Peak-layer accuracy on test pairs using a linear probe on the dense representations. $\text{Acc}_{\text{orig}}$: $A \to B$ with gold label. $\text{Acc}_{\text{flip}}$: $B \to A$ with label inverted (hypernymy $\leftrightarrow$ hyponymy). 95\% confidence interval reported as half-width in subscript.}
    \label{tab:hyper-hypo}
\end{table}

\subsection{SAE Analysis}
\subsubsection{Top-$k$ Selection}
To examine whether a small, ranked subset of pooled SAE features carries the causal signals for semantic relation inference, we need to select how many features to intervene on: too few may miss relevant signal, while too many dilute interpretability. We first rank features per relation and layer on the training set by the absolute probe weight for that relation, and then sweep over $k$ to pick a budget. We fix the selection grid to $k\in \{32, 64, 128, 160, 192, 224, 256, 296, 320\}$ and use $k=327$ as reference. 
We choose the smallest grid value whose selection score reaches at least 90\% of the effect at $k=327$ (1\% of the SAE latent size). The selection score is defined as the mean absolute change in the target class logit of the fixed probe on the training split under an intervention. This modification can be either keep-only (retaining only the top-$k$ features) or remove-only (ablating the top-$k$ features), with everything else fixed. For a linear probe, both variants yield identical absolute logit changes, so we report a single selection score and use the same $k$ for both sufficiency and necessity experiments.

\begin{figure*}[h]
    \centering
    \includegraphics[width=0.325\textwidth]{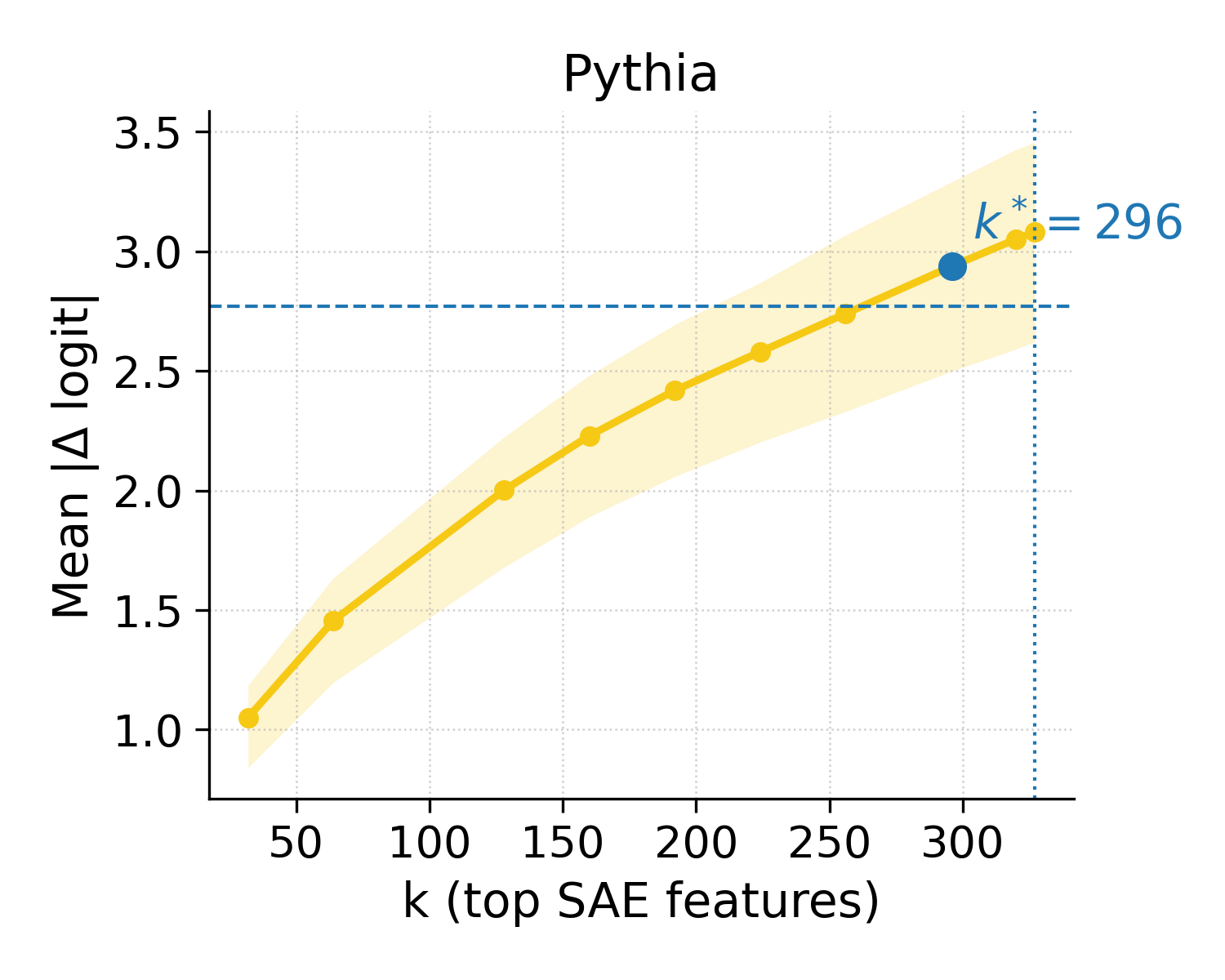}
    \hfill
    \includegraphics[width=0.325\textwidth]{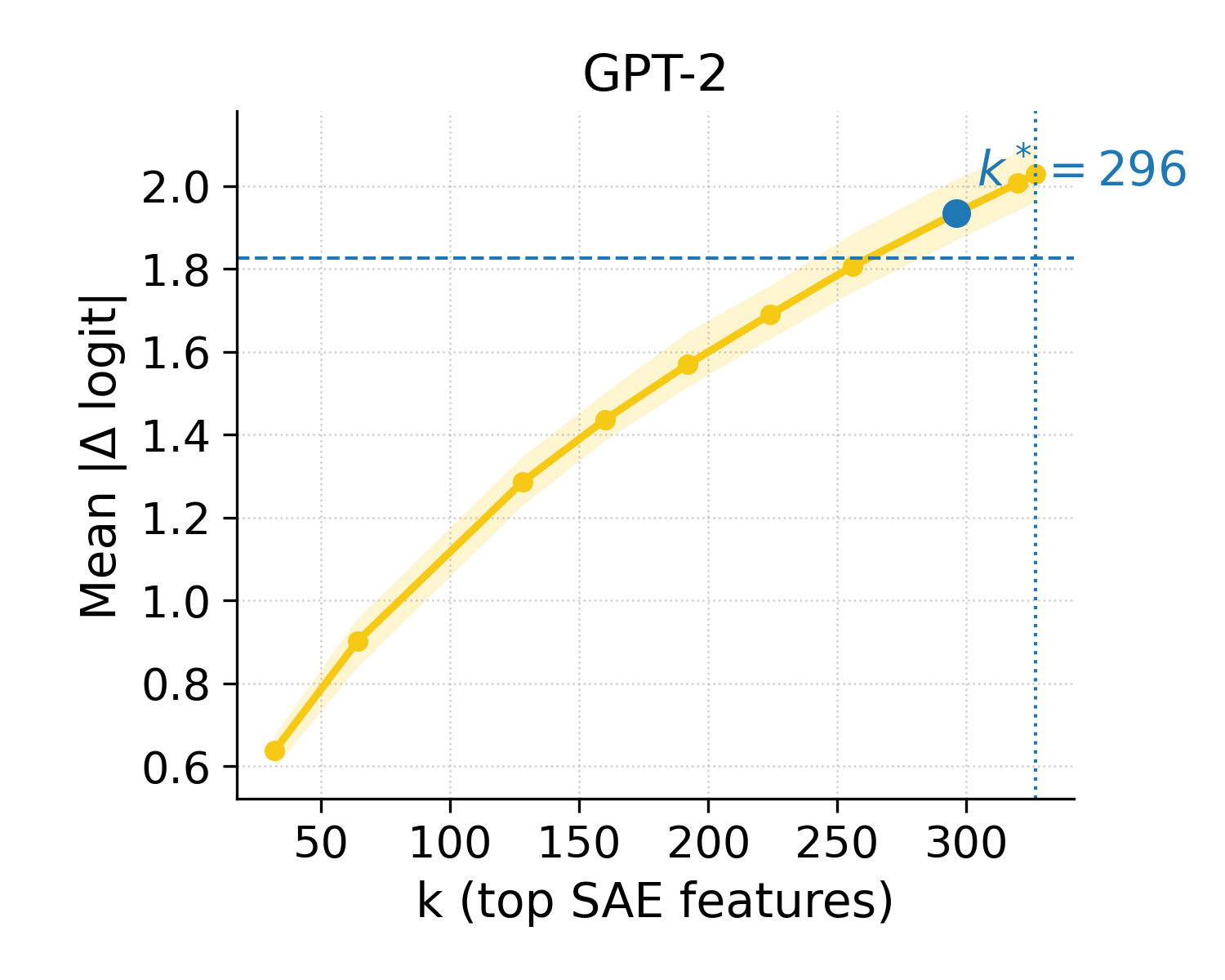}
    \hfill
    \includegraphics[width=0.325\textwidth]{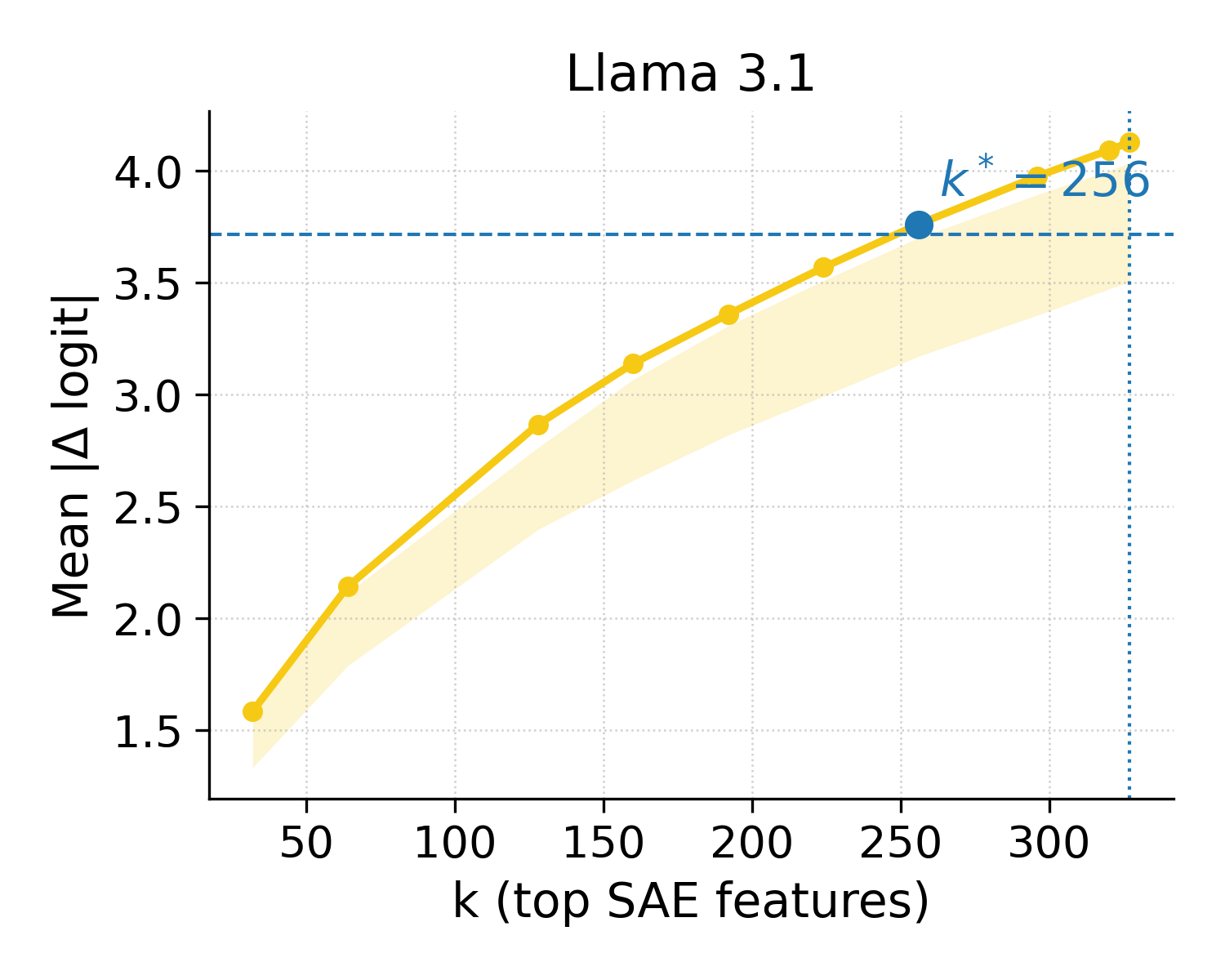}
    \caption{Mean change of target logit on the training set as a function of $k$. Yellow line: mean |$\Delta$ logit| over relations; shaded band: interquartile range across layers. Vertical dotted line: reference $k$ = 327 (1\% of SAE width). Horizontal dashed line: 90\% cutoff. Blue marker: smallest $k$ meeting the cutoff.}
    \label{fig:ksweep}
\end{figure*}

Figure~\ref{fig:ksweep} shows the results of the $k$-sweep. We select 296 for Pythia and GPT-2 and 256 for Llama 3.1 as $k$ value, and hold these fixed in all subsequent experiments on the test set.



\subsubsection{Activation Patching}
To test whether SAE features can causally drive the probe’s relational predictions, we perform necessity and sufficiency testing with activation patching on neutral inputs. 
Tables~\ref{tab:sufficiency-peak} and \ref{tab:necessity-peak} summarize standardized the semantic margin changes ($\Delta \text{LD}_{\text{sem}}$) and behavioral shifts ($\Delta$FR, DR) at each model’s peak layer (defined by the maximal $\Delta \text{LD}_{\text{sem}}$). 

Across models, only Llama 3.1 exhibits strong causal effects in both sufficiency ($\Delta \mathrm{FR} \ge 0.66$) and necessity ($\mathrm{DR} \ge 0.24$), with large positive and negative standardized margin shifts. Effects on GPT-2 are minimal ($\Delta \mathrm{FR} \le 0.09$; $\mathrm{DR} \le 0.14$), while on Pythia, shifts are slightly larger but remain unstable ($\Delta \mathrm{FR} \le 0.25$; $\mathrm{DR} \le 0.21$). As a control, we repeated each intervention with randomly sampled features of equal size $k$ (averaged over 5 seeds). Across all models and layers, random patching produced effects below $10\%$ of the corresponding top-$k$ effect, confirming that the observed shifts are specific to probe-ranked features rather than an artifact of modifying any $k$ latents.

Across relations, causal strength varies in both magnitude and consistency. Antonymy shows the largest behavioral shifts ($\Delta \mathrm{FR} = 0.79$; $\mathrm{DR} = 0.62$), while synonymy remains generally the weakest, mirroring the earlier probing test accuracies. Hypernymy shows high sufficiency but low necessity (i.\,e., ($\Delta \mathrm{FR} = 0.69$) vs $\mathrm{DR} = 0.35$ in Llama 3.1), suggesting diffuse or redundant coding. Hyponymy is more consistently high on both, indicating reliance on compact, targeted features. This asymmetry mirrors the directional bias observed in the word-order reversal experiments (Section \ref{sec:hyper-hypo}): hypernymy accuracy dropped by only $0.07-0.09$ under reversal, while hyponymy dropped by $0.24-0.29$ across all models. The SAE results offer a mechanistic account for this pattern: hypernymy is encoded redundantly across many features (high sufficiency, low necessity), making it robust to perturbation, whereas hyponymy relies on a compact set of features that are both sufficient and necessary, leaving it vulnerable when disrupted.
  

\begin{table}[t]
    \centering    
    \begin{adjustbox}{width=0.99\linewidth}
    \begin{tabular}{l|cc|cc|cc}
    \toprule
    & \multicolumn{2}{c|}{Pythia}
    & \multicolumn{2}{c|}{GPT-2}
    & \multicolumn{2}{c}{Llama 3.1} \\
    Relation
    & $\mathrm{std}\,\Delta \text{LD}_{\text{sem}}$ & $\Delta\text{FR}$
    & $\mathrm{std}\,\Delta \text{LD}_{\text{sem}}$ & $\Delta\text{FR}$
    & $\mathrm{std}\,\Delta \text{LD}_{\text{sem}}$ & $\Delta\text{FR}$ \\
    \midrule
    Syn   & 0.40$_{0.05}$ & 0.19 & 0.08$_{0.01}$ & 0.02 & 2.14$_{0.54}$ & 0.66 \\
    Ant   & 0.17$_{0.15}$ & 0.16 & 0.05$_{0.02}$ & 0.03 & 1.88$_{0.43}$ & \textbf{0.79} \\
    Hyper  & \textbf{0.49}$_{0.05}$ & \textbf{0.25} & \textbf{0.20}$_{0.01}$ & \textbf{0.09} & 3.07$_{0.53}$ & 0.69 \\
    Hypo   & 0.34$_{0.10}$ & 0.16 & 0.08$_{0.01}$ & 0.04 & \textbf{2.97}$_{0.75}$ & 0.75 \\
    \bottomrule
    \end{tabular}
    \end{adjustbox}
    \caption{Sufficiency results across models. Standardized $\Delta \text{LD}_{\text{sem}}$ (higher = stronger sufficiency) and the net flip-rate change toward target relation ($\Delta\text{FR}$) at peak layers. 95\% confidence interval reported as half-width in subscript.}
    \label{tab:sufficiency-peak}
\end{table}

\begin{table}[t]
    \centering
    \begin{adjustbox}{width=0.99\linewidth}
    \begin{tabular}{l
                    |cc
                    |cc
                    |cc}
    \toprule
    & \multicolumn{2}{c|}{Pythia}
    & \multicolumn{2}{c|}{GPT-2}
    & \multicolumn{2}{c}{Llama 3.1} \\
    Relation
    & $\mathrm{std}\,\Delta \text{LD}_{\text{sem}}$ & DR
    & $\mathrm{std}\,\Delta \text{LD}_{\text{sem}}$ & DR
    & $\mathrm{std}\,\Delta \text{LD}_{\text{sem}}$ & DR \\
    \midrule
    Syn   & -0.19$_{0.04}$ & 0.14 & -0.09$_{0.01}$ & 0.05 & -0.35$_{0.03}$ & 0.24 \\
    Ant   & -0.28$_{0.06}$ & \textbf{0.21} & \textbf{-0.31}$_{0.06}$ & 0.13 & -0.49$_{0.07}$ & \textbf{0.62} \\
    Hyper & -0.10$_{0.10}$ & 0.14 & -0.08$_{0.02}$ & 0.04 & -0.18$_{0.02}$ & 0.35 \\
    Hypo & \textbf{-0.36}$_{0.08}$ & 0.20 & -0.29$_{0.02}$ & \textbf{0.14} & \textbf{-0.94}$_{0.06}$ & 0.41 \\
    \bottomrule
    \end{tabular}
    \end{adjustbox}
    \caption{Necessity results across models. Standardized $\Delta \text{LD}_{\text{sem}}$ (lower = stronger necessity) and drop rate of target class prediction (DR) at peak layers. 95\% confidence interval reported as half-width in subscript.}
    \label{tab:necessity-peak}
\end{table}


\subsubsection{Contextual robustness}

To assess whether relation-specific representations depend on prompt phrasing or reflect context-independent structure, we evaluate the Llama 3.1 model on two unseen input formats: (a) a novel neutral template distinct from those used in training (\textit{``A occurs with B''}), and (b) a minimal format presenting only the word pair without any contextual sentence (\textit{``A B''}). Probes and activation patching are applied identically to those in earlier experiments.

The results in Table~\ref{tab:contextual_robustness} show that under prompt shifts, linear decodability degrades primarily in terms of mean accuracy ($-25\%$ and~$-17\%$), while peak accuracy remains largely stable across contexts. 
Causal sufficiency drops by $0.24$ in $\Delta\mathrm{FR}$ on the novel context and by $0.02$ on the no context format. Necessity increases by $0.16$ and $0.13$ in $\mathrm{DR}$.
Overall, our contextual robustness experiments show that relation signals appear mostly prompt-invariant in the representation space; the loss is in readout calibration rather than in feature availability.

\begin{table}[ht]
\small
\centering
    \begin{adjustbox}{width=0.48\textwidth}
    \begin{tabular}{l|cccc}
    \toprule
    Prompt Set & Mean & Peak & $\Delta\text{FR}$ & $\mathrm{DR}$ \\
    \midrule
    Original & 0.72$_{0.02}$ & 0.84$_{0.02}$ & 0.72 & 0.40 \\
    Novel context & 0.47$_{0.03}$ & 0.92$_{0.03}$ & 0.48 & 0.56 \\
    No context & 0.55$_{0.04}$ &  0.74$_{0.05}$ & 0.70 & 0.53 \\
    \bottomrule
    \end{tabular}
    \end{adjustbox}
    \caption{Contextual robustness of LLaMA-3.1 across unseen prompt templates, averaged across relations. Mean and Peak refer to test accuracy across layers. 
    95\% confidence interval reported as half-width in subscript.}    \label{tab:contextual_robustness}
\end{table}

\section{Conclusion}

We investigated if large language models encode explicit signals of semantic relations and how these signals are distributed across depth, components, and scale. We employed layer-wise linear probes with SAE-guided interventions to trace the origins of these signals and assess their manipulability.

Linear probes reveal consistent improvement in distinguishing relation types with model size and a stable difficulty ordering. The tasks rank from easiest to hardest: antonymy, hyponymy, hypernymy, and synonymy. 
Evidence concentrates in mid-layers, but the wide confidence intervals across depth indicate that relation information is not sharply localized to a single layer. Among subcomponents, MLP carries more recoverable evidence than attention, but neither alone matches the full block output. Word-order reversals expose a directional bias: hyponymy weakens under reversal, while hypernymy remains stable. At the same time, cosine similarity obscures relational distinctions even for symmetric pairs (synonymy, antonymy). Together, these findings indicate that semantic relations are encoded as structured, role-sensitive 
features rather than mere similarity.

To probe causality, we use sparse autoencoders to identify a small, ranked subset of features and intervene via targeted injection and ablation. These interventions yield strong, reliable effects on Llama 3.1, but are weaker and less stable on the smaller models. The activation patching outcomes reproduce the dense-probe difficulty ranking and surface an additional hyponymy–hypernymy asymmetry: hypernymy exhibits low necessity, while hyponymy is more strongly suppressed by ablations.

While our claims are limited to probe-level evidence, they nonetheless expose a consistent, manipulable structure in how models encode relations. From a practical standpoint, these findings could inform targeted interventions for tasks that depend on relational knowledge, such as ontology completion, textual entailment, or taxonomy-aware retrieval, where understanding which features encode specific relations and how robustly may guide more efficient fine-tuning or steering strategies. Future work may develop token-level interventions and evaluate their impact on generation, linking relation-specific features to next-token predictions and full decoded outputs.

\newpage
\section{Acknowledgments}

This research is co-funded by the CodeInspector project (No. 504226141) of the DFG, German Research Foundation.

\section{Ethical Considerations and Limitations}

This study focuses on the interpretability of large language models rather than their deployment in applied systems. Despite its analytical focus, this work entails various ethical and methodological concerns. First,  interpretability methods such as probing, sparse autoencoders, and activation patching yield partial, model-specific insights. Probes reveal correlations rather than true causality, and patching can exaggerate causal effects under nonlinear feature interactions. Furthermore, we operate at probe level rather than end-to-end behavior; links to model outputs remain indirect. Second, our model selection spans three scales but is limited to base, dense decoder-only architectures. Instruction-finetuned variants, mixture-of-experts models, and encoder-only architectures may encode semantic relations differently, and we leave such comparisons to future work. Third, the use of WordNet as ground truth introduces representational bias, as its English-centric lexical structure may not generalize across languages or domains. Finally, sparse feature decomposition methods such as SAEs raise reproducibility and privacy considerations. Disentangled features may expose memorized or identifiable patterns, a risk mitigated by relying on open-weight models and restricting analyses to WordNet-based lexical inputs.
\clearpage
\section{Bibliographical References}\label{sec:reference}

\bibliographystyle{lrec2026-natbib}
\bibliography{lrec2026-example}

\section{Language Resource References}
\label{lr:ref}
\bibliographystylelanguageresource{lrec2026-natbib}
\bibliographylanguageresource{languageresource}

\end{document}